\documentclass[runningheads]{llncs}
\usepackage[T1]{fontenc}
\usepackage{graphicx}
\usepackage{booktabs}
\usepackage[misc]{ifsym}
\newcommand{\corr}{(\Letter)}
\usepackage[numbers,square]{natbib}
\usepackage{tikz}
\usepackage{float}

\usepackage{pgfplots}
\usepackage{pgfplotstable}
\pgfplotsset{compat=1.18}
\usepackage{filecontents}
\usepackage{multirow}
\usepackage{amsmath}
\usepackage[capitalise]{cleveref}

\newcommand{\tf}{\text{tf}}
\newcommand{\W}{\text{W}}
\newcommand{\mean}{\mu}

\begin{document}

\title{Enhancing Traffic Accident Classifications: Application of NLP Methods for City Safety}

\titlerunning{Enhancing Traffic Accident Classifications}
% If the full title of your paper is short enough to also fit in the running head, you can omit the abbreviated paper title here. You can check as follows: if you comment out the \titlerunning line, something will appear in the header of all odd-numbered pages of your PDF from page 3 onward. This something is either the full title (in which case all is well), or the error message "Title Suppressed Due to Excessive Length". If this error message appears, you're going to want to provide an abbreviated title within the \titlerunning command, because if you won't do it, Springer will do it for you.

%N.B.: Author information (both in the \author{} and \authorrunning{} command) should only be present in the Camera-Ready Version of your paper. The version that you initially submit for review, ought to be double-blind. So, when initially submitting your paper, use:
%\author{Author information scrubbed for double-blind reviewing}
% \author{Andr\'e Lauren Benjamin\inst{1} \and
% Calvin Cordozar Broadus Jr.\inst{2,3} \corr \and
% Antwan Andr\'e Patton\inst{1}\orcidID{0000-1111-2222-3333}}
% You may leave out the orcidID information, if you want to.
% Use \corr to indicate the corresponding author. Note the spacing around the \corr command. Only one author can be the corresponding author.
\author{Enes Özeren\inst{1}\thanks{These authors contributed equally to this work.} \and
Alexander Ulbrich\inst{1\ast} \and
Sascha Filimon\inst{2} \and
David Rügamer\inst{1, 3} \and
Andreas Bender\inst{1, 3} \corr
}

%ECML submission requirement
\tocauthor{Enes Özeren, Alexander Ulbrich, Sascha Filimon, David Rügamer, Andreas Bender}
\toctitle{Enhancing Traffic Accident Classifications: Application of NLP Methods for City Safety}

%N.B.: comment out the \authorrunning{} command for the double-blind version of your paper submitted for review. Later, if your paper is accepted, use the command for the Camera-Ready Version.
% \authorrunning{A.L. Benjamin et al.}
\authorrunning{E. Özeren and A. Ulbrich et al.}
% If there are two authors, write 'A.L. Benjamin and C.C. Broadus Jr.'
% If there are more than two authors, '[...] et al.' is used.

\institute{Department of Statistics, LMU Munich, Munich, Germany \email{\{enes.oezeren,a.ulbrich\}@campus.lmu.de}
\and
City of Munich, Munich, Germany
\and
Munich Center for Machine Learning (MCML), Munich, Germany \email{andreas.bender@stat.uni-muenchen.de}}
% \institute{Fictional Southern University, Savannah GA 31404, USA \email{\{a.l.benjamin,a.a.patton\}@fsu.fake}
% \and
% Fictional West Coast University, Long Beach CA 90840, USA \email{ccb@fwcu.fake}
% \and
% Secondary European Affiliation, Tiergartenstr. 17, 69121 Heidelberg, Germany
% \email{lncs@springer.com}}

\maketitle              % typeset the header of the contribution

\begin{abstract}
%The abstract should briefly summarize the contents of the paper in
A comprehensive understanding of traffic accidents is essential for improving city safety and informing policy decisions. In this study, we analyze traffic incidents in Munich to identify patterns and characteristics that distinguish different types of accidents. The dataset consists of both structured tabular features, such as location, time, and weather conditions, as well as unstructured free-text descriptions detailing the circumstances of each accident. Each incident is categorized into one of seven predefined classes. To assess the reliability of these labels, we apply NLP methods, including topic modeling and few-shot learning, which reveal inconsistencies in the labeling process. These findings highlight potential ambiguities in accident classification and motivate a refined predictive approach. Building on these insights, we develop a classification model that achieves high accuracy in assigning accidents to their respective categories. Our results demonstrate that textual descriptions contain the most informative features for classification, while the inclusion of tabular data provides only marginal improvements. These findings emphasize the critical role of free-text data in accident analysis and highlight the potential of transformer-based models in improving classification reliability.
\end{abstract}

\section{Introduction}
Traffic accidents pose a substantial risk to human life and incur high economic costs. Understanding their underlying causes and patterns is essential - not only to mitigate their consequences but also to develop effective preemptive strategies in the context of city safety and city planning. Numerous studies have investigated various aspects of traffic accidents, from identifying their root causes to assessing their severity using data analysis techniques \citep{hossain2023data, park2021scenario, seo2023text, wu1998natural}. The foundation of such studies is the availability of high-quality accident data. However, real-world accident records are often affected by inconsistencies and human error during data collection. These issues can lead to inaccuracies in how accidents are categorized, potentially obscuring important insights and limiting the effectiveness of data-driven safety measures.

\begin{table}[h]
    \centering
        \caption{Classification of accidents and their explanations.}
    \label{tab:accident_classification}
\resizebox{\columnwidth}{!}{    \begin{tabular}{clp{0.2cm}p{10cm}}
        % \hline
        \textbf{Code} & \textbf{Classification} && \textbf{Explanation} \\
        \hline
        A1 & Driving Accident && Loss of control of vehicle. \\
        %\hline
        A2 & Turning / Crossing Accid.\ && Conflict between turning vehicle and one moving in parallel direction. \\
        %\hline
        A3 & Turning Accident && Conflict betw.\ (turning) vehicle and another moving perpendicularly. \\
        %\hline
        A4 & Crossing Accident && Conflict between vehicle and crossing pedestrian. \\
        %\hline
        A5 & Stationary Accident && Conflict where at least one party must be stationary/parking. \\
        %\hline
        A6 & Longitudinal Accident && Conflict between parties moving in parallel, none of above applicable. \\
        %\hline
        A7 & Other Accident && None of the above applicable. %\\
        %\hline
    \end{tabular}
    }
\end{table}

In the city of Munich, Germany, policemen record accident information at the location of the incident, which includes general information like date, time and location, person-specific characteristics (age, drug involvement, injury severity) as well as free-text description of the events leading up to the incident. The specific dataset used in this study is comprised of 105,217 unique traffic accidents recorded between 01.01.2017 and 31.12.2022, of which 102,569 contain free-text.
Additionally, on-site, the policemen classify the accidents into one of seven distinct accident types (A1 -- A7), listed in \cref{tab:accident_classification}. This is done by (mentally) matching the course of events to the definition of the respective accident types. In order to avoid confusion later on, we define the following terms:

\begin{itemize}
\item \emph{label definition}: A textual definition for each of the seven accident types (A1 - A7).
\item \emph{example text} or \emph{accident description}: The free-text description of the accident recorded by the policemen on-site.
\item \emph{human label}: The label (A1 - A7) assigned to an accident on-site by the policemen.
\item \emph{ground truth}: 236 additional labels created by expert labelers for a small subset of accidents.
\end{itemize}

In our data set, almost 50\% of the human labels fall within the fallback category A7. Given the high proportion of accidents in this category, a high misclassification rate is suspected. This is further supported by comparison to the city of Berlin, where only 25\% of accidents fall into this fallback category. While there may be inherent differences between the two cities, the large proportion of accidents in the fallback category (A7) indicates a potentially high amount of mislabeling.

In this work, we aim to gain insights into the reasons for mislabeling and to improve the current classification system. This could lead to the implementation of better and more accurate safety measures. To this end, we utilize multimodal data, comprising free-text incident descriptions as well as tabular data.

Using advanced Natural Language Processing (NLP) techniques that have been successfully applied in various domains, including medical records, legal documents, and police reports \citep{li2024scoping,siino2025exploring,xing2024entity}, we gain insights about missclassification by applying transformer-based methods. Furthermore, we develop a classification model that achieves high accuracy in correctly assigning accidents. We make the code publicly available.\footnote{\url{https://github.com/enesozeren/enhancing-traffic-accident-classifications}}

\section{Related Work}

\subsection{Traffic Accident Analysis}

With the growing popularity of NLP methods, recent research has increasingly explored their application in accident data analysis \citep{park2021scenario, seo2023text, wu1998natural}. While structured data has been extensively studied (see, e.g., \citep{hossain2023data, nassereddine2024evaluating}), the use of unstructured free-text features has gained traction only in recent years. Free-text descriptions can encode nuanced information that structured numerical data cannot fully capture, offering deeper contextual insights \citep{wu1998natural}.
Early approaches to leveraging free-text descriptions often relied on simple word-count-based methods, such as \citep{park2021scenario}, where text features were extracted based on keyword frequencies. Beyond traffic accidents, similar methods have been applied in legal text analysis. For instance, \citep{noguti2020legal} achieved strong results using LSTMs for petition analysis and suggested exploring transformer-based methods as a next step.
Recent research has increasingly focused on using word embeddings to capture richer semantic information. In this context, \citep{seo2023text} employ BERT to incorporate free-text accident descriptions, demonstrating promising performance in a classification setting. Their findings suggest that further integrating free-text features could unlock significant potential, as such descriptions are widely used across different countries \cite{seo2023text}. Their work focuses on extracting specific information from the accident descriptions and to compare classical and modern NLP approaches for classification, assuming the human labels to represent ground truth. In contrast, we investigate potential mislabeling of accident types and use multiple data modalities for classification compared to text-based inputs only.

\subsection{Large Language Models}

The Transformer architecture, proposed in 2017 for machine translation tasks \citep{vaswani2017attention}, quickly became the dominant paradigm in the NLP domain \citep{patwardhan2023transformers}. The original design consists of an encoder-decoder structure, but both components are also independently used. While encoder-only models are utilized primarily for learning text representations \citep{conneau2019unsupervised,devlin2019bert}, encoder-decoder and decoder-only models are employed for text generation tasks \citep{radford2018improving, raffel2020exploring}.

These models, often referred to as Large Language Models (LLMs), have a large number of parameters and are trained on massive text corpora \citep{brown2020language, conneau2019unsupervised, devlin2019bert, radford2018improving, team2024gemma}. They can be fine-tuned for specific tasks, allowing for domain adaptation and improved performance \citep{devlin2019bert, radford2018improving}. Alternatively, techniques such as few-shot and chain-of-thought prompting have enabled the application of these models with good performance without requiring any additional parameter updating \citep{brown2020language, wei2022chain}. In this project, we utilized both approaches, few-shot classification for creating a second opinion about accident categories, and also fine-tuning for predictive modeling.

\subsection{Topic Modeling}

Topic models are designed to extract semantic themes from large volumes of unstructured text \citep{blei2012probabilistic}. Traditional approaches such as latent dirichlet allocation (LDA) and non-negative matrix factorization (NMF) have been widely used for topic modeling. However, their performance is limited by a lack of semantic understanding, as they rely solely on bag-of-words representations and fail to capture contextual information \citep{grootendorst2022bertopic}.
Therefore, novel methods incorporating text-embeddings have been increasingly applied to topic modeling. In \citep{grootendorst2022bertopic}, BERTopic is proposed, a framework to perform topic modeling by creating dense vector representations of each document, which are then used for clustering. The framework makes use of Sentence-BERT \citep{reimers-gurevych-2019-sentence}, a time-efficient alternative of BERT \citep{devlin2019bert} enabling to compare embeddings with cosine similarity. The resulting embeddings are dimensionally reduced and clustered. Finally, text representations for each cluster are chosen by modifying the TF-IDF approach proposed by \citep{joachims1997probabilistic} in a way that all documents within one cluster are treated as a single document \citep{grootendorst2022bertopic}. This allows to extract class-related representative keywords. In our study, topic modeling is used to identify relevant topics within misclassified accidents.

\section{Semantic Clustering}

To investigate potential mislabeling, we first evaluate what semantic characteristics the free-text descriptions of accidents in a certain category show. This enables us to discover patterns within each accident type, notably which topics show up frequently within the fallback category A7. 

\subsection{Methods}
In order to perform semantic clustering, we apply BERTopic \citep{grootendorst2022bertopic} to extract topics present in the text corpus. Clustering is performed in an unsupervised way using the free-text accident descriptions only and without taking into account their human labels (A1-A7). 

The first step is to convert all texts into dense vector representations. While BERTopic allows for direct usage without specifying a specific model, it is also possible to encode the text independently and pass the resulting vectors as an additional argument. One requirement for a suitable Sentence-Transformer is a context window of at least 2000 tokens, as this is the maximal text length in the dataset. Furthermore, the model is required to have German capabilities. For the study and given the two requirements, jiina-embeddings-v3 \citep{sturua2024jinaembeddingsv3multilingualembeddingstask} is chosen from the MTEB benchmark ranking \citep{muennighoff2022mteb}. The model performs mean-pooling by default for combining all token-vectors into a single vector for each text.

Since the resulting embedding-vectors are 1024-dimensional, their dimensionality can be reduced. UMAP has shown to be able to reduce the amount of dimensions while maintaining more of the global structure than competing methods like t-SNE or PCA \citep{mcinnes2018umap}. Four hyper-parameters have to be specified when using UMAP \citep{mcinnes2018umap}. \textit{Number of dimensions} controls the number of dimensions the reduced vector should have. \textit{Number of neighbors} influences the locality of approximation patterns. If the parameter is increased, more global structures will be captured. In the context of this study, if one would be interested in many fine-grained topics, \textit{number of neighbors} can be decreased. \textit{Minimal distance} is mainly important for plotting since it controls how densely points can be packed together. It can be increased to avoid overplotting.

Next, the reduced embedding vectors can be clustered. For the study HDBSCAN is used, an extension of DBSCAN which allows for capturing clusters with varying densities \citep{mcinnes2017hdbscan}. As a hyper-parameter, a minimal cluster size can be fixed.  
Finally, for each cluster, representations have to be generated. The goal is to find words which are relevant for certain clusters. In \citep{grootendorst2022bertopic} it is proposed to employ a variation of TF-IDF, such that documents within each cluster are considered as single documents, giving rise to the c-TF-IDF approach.
\begin{equation}
\W_{t,c} = \tf_{t,c} \cdot \log\left(1 + \frac{\mean}{\tf_t} \right)
\end{equation}

Here the term frequency $\tf_{t,c}$ indicates the frequency of word $t$ in cluster $c$. $\mean$ is the average number of words per class while $\tf_t$ is the frequency of term $t$ across all classes. $\W_{t,c}$ can therefore be interpreted as an estimated importance score for word $t$ within class $c$. It needs to be stressed that this formula does not take into account word embeddings, but is only based on word frequencies. Therefore it might fail to accurately capture the true semantic meaning of each extracted topic \citep{grootendorst2022bertopic}.

\subsection{Results}

Due to computational constraints, we limited our analysis to a random subset of 50,000 accident descriptions.
The topic model extracts 18 different topics, listed in Table~\ref{tab:topic-modeling}. It can be seen that multiple topics about parking accidents have been extracted. Despite looking similar according to representative documents and the selected c-TF-IDF representations, different nuances are captured within some of those topics. To give one example, ``Parking 3'' has a relatively high cosine similarity to the topic ``Intox.'' which is about accidents related to drug influence. Considerations like this can give an idea of what different subtleties the seemingly identical topics show.

\setlength{\tabcolsep}{2pt}
\begin{table}[!ht]
\centering
\caption{Topics extracted by BERTopic (outliers excluded). Topic (column 1) shows subjectively labeled topic names, based on representative documents and the extracted c-TF-IDF terms for better readability. Topics are ordered in descending order with regards to their counts (column 2), i.e., the number of observations per topic. The third column includes the content of each topic.}
\label{tab:topic-modeling}
\begin{tabular}{lll}
\toprule
Topic & Count & Content \\
\midrule
Parking 1 & 13,846 & Parking accidents \\
Bicycle & 8,293 & Bike accidents, falling from bikes \\
Crossroad/Crash & 4,376 & Accidents mostly in crossroads, many with crashes \\
Parking 2 & 2,374 & Parking accidents \\
Parking 3 & 2,212 & Parking accidents \\
Parking 4 & 1,457 & Parking accidents \\
Truck & 1,242 & Truck accidents, mostly damaging parked vehicles \\
Parking 5 & 999 & Parking accidents, damaged side mirrors \\
Bus & 828 & Bus accidents \\
Damaged city obj. & 645 & Damaged objects like traffic lights, fences or traffic signs \\
Scooter & 540 & Scooter- and motorcycle accidents \\
Landsbergerstr. & 422 & Accidents in spacial proximity to Landsbergerstreet \\
Schleißheimerstr. & 418 & Accidents in spacial proximity to Schleißheimerstreet \\
Intox. & 399 & Accidents connected to drug influence \\
Fürstenriederstr. & 394 & Accidents in spacial proximity to Fürstenriederstreet \\
Dachauerstr. & 366 & Accidents in spacial proximity to Dachauerstreet \\
Parking 6 & 305 & Parking accidents \\
Tram & 301 & Tram accidents \\
\bottomrule
\end{tabular}
\end{table}

\begin{figure}[t]
    \centering
    \includegraphics[width=0.8\textwidth]{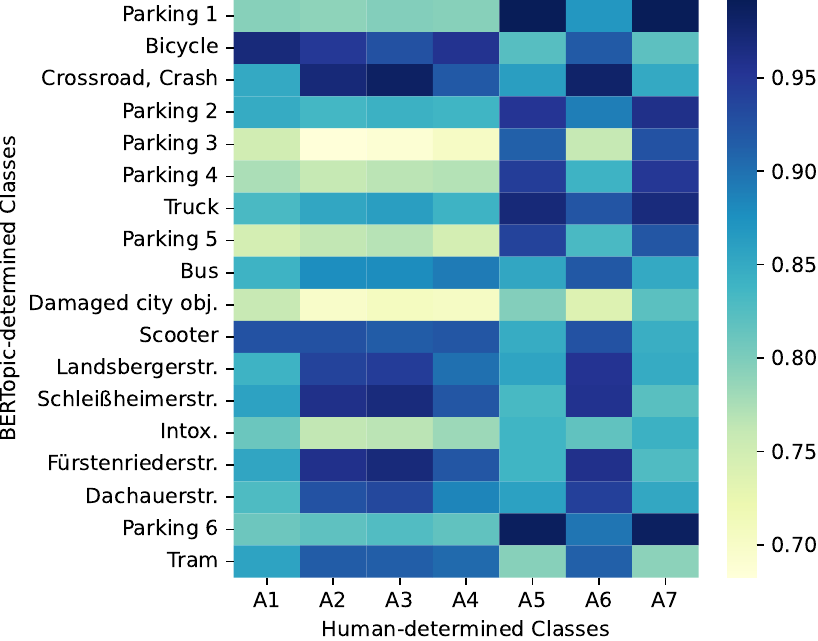}
    \caption{Cosine similarity between BERTopic-generated and human-determined clusters.}
    \label{fig:bertopic-human-similarites}
\end{figure}

As one objective of the study is to understand what kind of accidents tend to get the fallback label A7 (other accident), the resulting topics can now be compared to the human labels. To do this, the text corpus is clustered in two different ways:
\begin{enumerate}
    \item The texts are divided as suggested by the topic model, giving rise to 18 clusters.
    \item The texts are divided as suggested by the human labels, i.e., the class assignment to one of the 7 categories is used as cluster indicator. This yields 7 clusters (A1 -- A7).
\end{enumerate}
Both clustering schemes are used in the following way. After encoding all texts with the same model used for clustering, representative embedding vectors are generated by applying mean-pooling within each cluster defined above. After this, their cosine similarity can be calculated and summarized as depicted in Figure~\ref{fig:bertopic-human-similarites}.

First, we note that the lowest cosine similarity value is around 0.7, indicating generally high similarity, as this measure ranges up to a maximum of 1. This might be due to the fact that all texts are similar in the way that they all deal with traffic accidents. What can also be seen, for instance, is a high similarity between accident type A1 (driving accident), and the bicycle topic from the topic model (column 1). In fact, this accident type represents the class among all 7 with the highest proportion of bikes involved. Looking at column five, which includes accident type A5 (stationary accidents), a relatively high similarity to all six extracted parking topics is visible. Finally, column seven, which represents the fallback category A7 (other accidents) shows almost exactly the same color pattern as column five. In other words, accidents labeled as ``other accident'' seem to be similar to what our topic model identifies as parking accidents.

%%%%%%%%%%%%%%%%%%%%%%%%%%%%%%%%%%%%%%%%%%%%% Few Shot %%%%%%%%%%%%%%%%%%%%%%%%%%%%%%%%%%%%%%%%%%%%%%
\section{Classification with Few-Shot Prompting}
\label{sec:few-shot}
In this section, labels are generated based on few-shot prompting techniques, designed to mimic the human labeling process in that it matches an accident description to the label definition of an accident type. The results are compared to the human labels, revealing potential anomalies in the labeling behavior.

\subsection{Methods}
Few-shot prompting is performed by conditioning the LLM on a given task description and a small set of examples to solve a task \citep{brown2020language}. This approach has been proven to work better than zero-shot prompting, which relies only on task description without examples \citep{brown2020language}. Unlike fine-tuning, few-shot prompting does not involve updating model parameters and a small number of examples (typically 2-10) is sufficient for effective task adaptation. This makes it more efficient than fine-tuning, which generally requires hundreds or even thousands of labeled examples for language tasks. 

To apply few-shot prompting for our accident classification, a suitable LLM is selected based on three requirements. First, the model should be open-source to ensure it can process confidential data locally. Second, it needs to have strong proficiency in German, as all our text data is in German. Lastly, a technical requirement is that the model should run on available hardware (two Nvidia RTX A6000 GPUs with 48GB memory in each). Based on these criteria, we choose the Gemma-2-27B-Instruct model by Google \citep{team2024gemma} as it meets our requirements and demonstrated strong performance in five widely used German language benchmarks \citep{thellmann2024towards}.

\subsection{Results}
\label{ssec:results-topics}
\begin{figure}[!ht]
\includegraphics[width=1\textwidth]{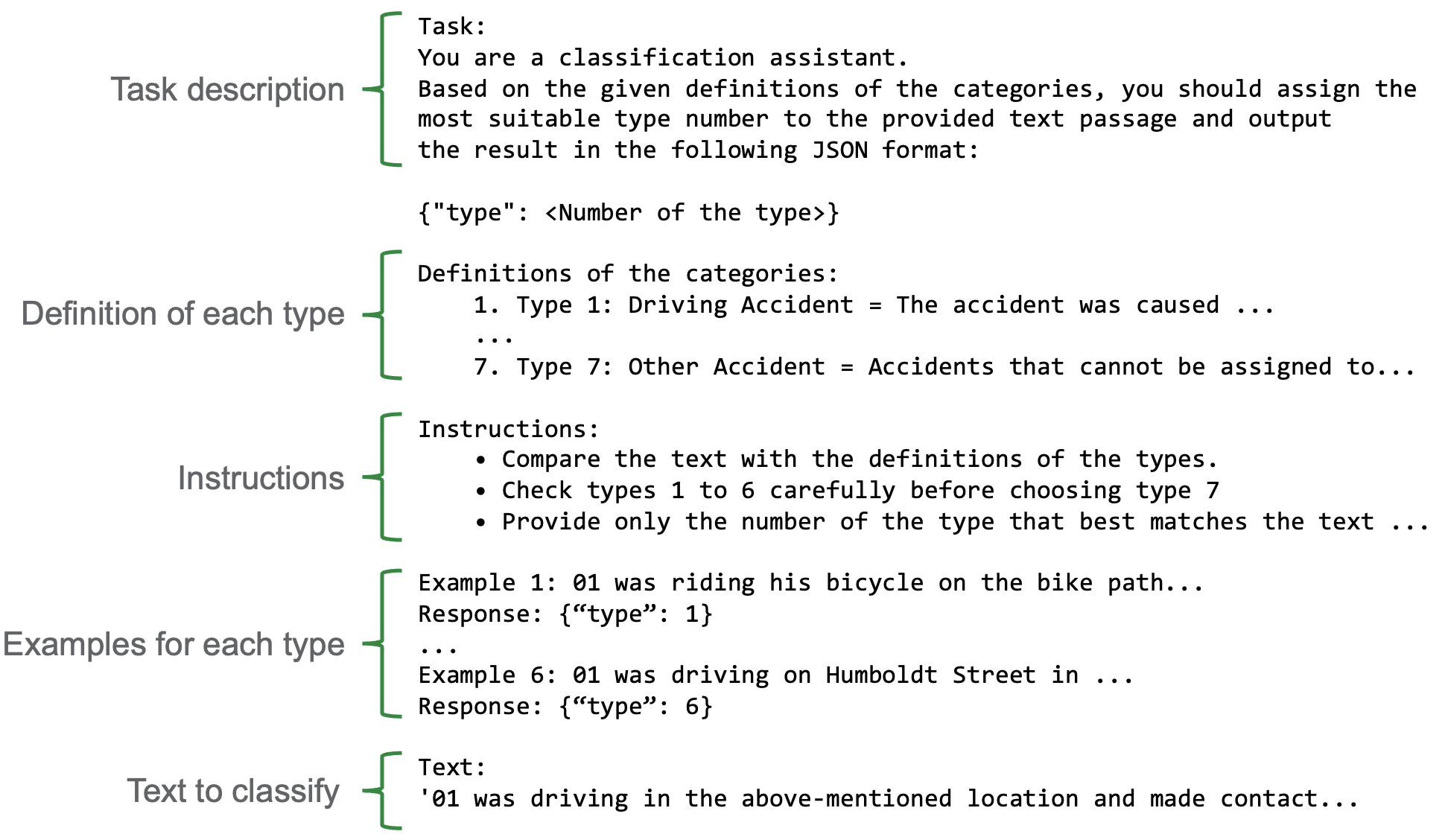}
\caption{Few-shot prompt for classifying accidents. Some components of the prompt shortened for illustration. The last part (text to classify) is changed for each accident text and inference is performed with the Gemma-2-27B-Instruct model. This is the translated English version; the original prompt is in German since accident texts are also in German.} 
\label{fig:few-shot-prompt}
\end{figure}

For our analyses, we use the label definitions that provide a high-level, representative description of each accident type and select six exemplary accident descriptions (with verified labels) for each non-fallback accident type (categories A1 to A6) from our data set. We intentionally exclude an example for accident type A7, which is the fallback category, to prevent the model from becoming biased towards a specific instance of category A7 (e.g., an accident involving a deer). We also refrained from including multiple examples per accident type to maintain a manageable prompt length, given hardware constraints (two Nvidia RTX A6000 GPUs). This decision was necessary to keep the inference time feasible for classifying all the accidents in the dataset, which already required approximately 24 hours with our current setup. The (shortened) prompt is given in Figure \ref{fig:few-shot-prompt}.

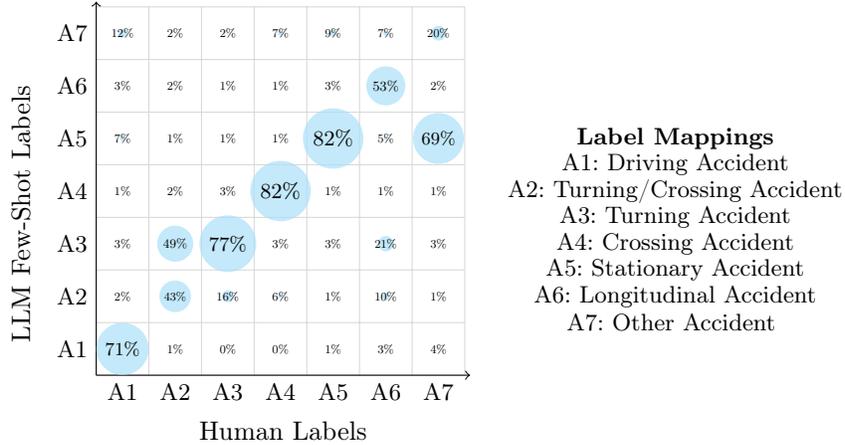
\begin{figure}
\centering
\begin{tikzpicture}[scale=0.70]
% Set up the grid and axes
\draw[gray!30, step=1] (0,0) grid (7,7);
\draw[->] (0,0) -- (7.1,0) node[pos=0.5, below=0.5cm] {\normalsize Human Labels};
\draw[->] (0,0) -- (0,7.1) node[pos=0.8, left=1cm, rotate=90] {\normalsize LLM Few-Shot Labels};

% Add axis labels
\foreach \x [count=\i from 1] in {A1,A2,A3,A4,A5,A6,A7} {
    \node[below] at (\i-0.5,0) {\small \x};
}
\foreach \y [count=\j from 1] in {A1,A2,A3,A4,A5,A6,A7} {
    \node[left] at (0,\j-0.5) {\small \y};
}

% Define bubble data - manually specify values
\foreach \x/\y/\size/\text in {
    0.5/0.5/0.71/71\%, 0.5/1.5/0.02/2\%, 0.5/2.5/0.03/3\%, 0.5/3.5/0.01/1\%, 0.5/4.5/0.07/7\%, 0.5/5.5/0.03/3\%, 0.5/6.5/0.12/12\%,
    1.5/0.5/0.01/1\%, 1.5/1.5/0.43/43\%, 1.5/2.5/0.49/49\%, 1.5/3.5/0.02/2\%, 1.5/4.5/0.01/1\%, 1.5/5.5/0.02/2\%, 1.5/6.5/0.02/2\%,
    2.5/0.5/0.0/0\%, 2.5/1.5/0.16/16\%, 2.5/2.5/0.77/77\%, 2.5/3.5/0.03/3\%, 2.5/4.5/0.01/1\%, 2.5/5.5/0.01/1\%, 2.5/6.5/0.02/2\%,
    3.5/0.5/0.0/0\%, 3.5/1.5/0.06/6\%, 3.5/2.5/0.03/3\%, 3.5/3.5/0.82/82\%, 3.5/4.5/0.01/1\%, 3.5/5.5/0.01/1\%, 3.5/6.5/0.07/7\%,
    4.5/0.5/0.01/1\%, 4.5/1.5/0.01/1\%, 4.5/2.5/0.03/3\%, 4.5/3.5/0.01/1\%, 4.5/4.5/0.82/82\%, 4.5/5.5/0.03/3\%, 4.5/6.5/0.09/9\%,
    5.5/0.5/0.03/3\%, 5.5/1.5/0.10/10\%, 5.5/2.5/0.21/21\%, 5.5/3.5/0.01/1\%, 5.5/4.5/0.05/5\%, 5.5/5.5/0.53/53\%, 5.5/6.5/0.07/7\%,
    6.5/0.5/0.04/4\%, 6.5/1.5/0.01/1\%, 6.5/2.5/0.03/3\%, 6.5/3.5/0.01/1\%, 6.5/4.5/0.69/69\%, 6.5/5.5/0.02/2\%, 6.5/6.5/0.2/20\%
} {
    \fill[cyan!30, opacity=0.7] (\x,\y) circle ({0.7*\size});
    \node[scale={max(0.5, 1.1*\size)}] at (\x,\y) {\text};
}

% Legend
\begin{scope}[shift={(11,3.5)}]
    \node[align=center, font=\bfseries] at (0,1) {Label Mappings}; % Moved higher
    \foreach \y/\text in {
        0.5/A1: Driving Accident,
        0.0/A2: Turning{/}Crossing Accident,
        -0.5/A3: Turning Accident,
        -1.0/A4: Crossing Accident,
        -1.5/A5: Stationary Accident,
        -2.0/A6: Longitudinal Accident,
        -2.5/A7: Other Accident
    } {
        \node[align=left] at (0,\y) {\text};
    }
\end{scope}

\end{tikzpicture}
\caption{Distribution of LLM few-shot labels vs.\ human labels. Each column sums to 100\%, and bubble sizes correspond to the intersection ratio between LLM few-shot and human labels. The anti-diagonal represents the agreement ratio between the two.}
\label{fig:bubble-plot}
\end{figure}

We apply the few-shot prompting for each accident description individually using the Gemma model and compare the results with human labels, as shown in Figure \ref{fig:bubble-plot}. Overall, 44\% of the LLM few-shot labels match the human labels. The anti-diagonal indicates a high agreement ratio for most accident types, but the large bubbles outside of the diagonal serve as an important signal for deeper analysis.

There are three bubbles larger than 20\% outside the anti-diagonal in Figure \ref{fig:bubble-plot}. The first case represents 49\% of accidents labeled as type A2 by humans but classified as type A3 by LLM few-shot approach. Since A2 and A3 accident types are both variations of turning accidents with a small difference (the driving direction of vehicles), we observe that LLM confused them easily. Upon reviewing examples, we find that humans could distinguish these cases more accurately than the LLM few-shot approach, potentially because they have a physical view of the accident scene which is not represented accurately in the textual description. The second case represents 21\% of accidents labeled as type A6 (longitudinal accident) by humans but classified as type A3 (turning/crossing accident) by the LLM few-shot approach. Similar to the first case, human judgment is more reliable as they can interpret the temporal nature of events, whereas the LLM misclassifies those longitudinal accidents occurring just after turning maneuvers.

The third and most notable case is where 69\% of accidents labeled as A7 (other accidents) by humans are classified as A5 (stationary accident) by the LLM. This case accounts for 34\% of all traffic accidents. We observe that these accidents consistently contain parking-related keywords such as 'garage', 'parking', etc., as well as misspelled variations of them, indicating that they are related to damaged parked vehicles (which is in line with our findings from Section \ref{ssec:results-topics}). This finding helps reduce uncertainty about accident characteristics in the fallback category A7. Before, 49\% of all accidents in Munich fell into the fallback category, ``Other accident (A7)'', meaning their specific nature is unknown. Our analysis reveals that most of these accidents involve damages to parked vehicles. As a result, the proportion of accidents of unknown nature can be reduced substantially.
This has practical implications. Only accidents of a known nature can be counteracted by city planning. For example, if an accident of types A1-A6 occurs frequently within a specific time span and location, countermeasures (e.g., traffic signs, etc.) can be implemented. For the fallback-category, this is not the case, as accidents could be of heterogeneous nature. However, our analysis helps to identify a large proportion of those as parking related, which can be mitigated accordingly.

\section{Predictive Modeling}

In this section, we describe our approach to building predictive models for accident classification. Given the potential for mislabeling in the training data, we explore different strategies for constructing training sets to improve label quality. Since both tabular and text data are available, we investigate these modalities individually and in combination to assess their contributions to predictive performance. Finally, we evaluate the models on an expert-labeled, ground-truth test set to provide a reliable assessment of their accuracy.

\subsection{Methods}
\subsubsection{Evaluation strategy} 
\begin{figure}
    \centering
    \begin{tikzpicture}[scale=0.65]
    % Left Bubble Plot
    \begin{scope}[shift={(-20,0)}]
        % Grid and Axes
        \draw[gray!30, step=1] (0,0) grid (7,7);
        \draw[->] (0,0) -- (7.1,0) node[pos=0.5, below=0.5cm] {\normalsize Human Labels};
        \draw[->] (0,0) -- (0,7.1) node[pos=0.9, left=0.7cm, rotate=90] {\normalsize LLM Few-Shot Labels};

        % Axis Labels
        %\foreach \x in {1,...,7} {\node[below] at (\x-0.5,0) {\large \x};}
        %\foreach \y in {1,...,7} {\node[left] at (0,\y-0.5) {\large \y};}
        \foreach \x [count=\i from 1] in {A1,A2,A3,A4,A5,A6,A7} {
        \node[below] at (\i-0.5,0) {\small \x};
        }
        \foreach \y [count=\j from 1] in {A1,A2,A3,A4,A5,A6,A7} {
        \node[left] at (0,\j-0.5) {\small \y};
        }

        % Bubbles
        % Define bubble data - manually specify values
        \foreach \x/\y/\size/\text in {
            0.5/0.5/0.71/71\%, 0.5/1.5/0.02/2\%, 0.5/2.5/0.03/3\%, 0.5/3.5/0.01/1\%, 0.5/4.5/0.07/7\%, 0.5/5.5/0.03/3\%, 0.5/6.5/0.12/12\%,
            1.5/0.5/0.01/1\%, 1.5/1.5/0.43/43\%, 1.5/2.5/0.49/49\%, 1.5/3.5/0.02/2\%, 1.5/4.5/0.01/1\%, 1.5/5.5/0.02/2\%, 1.5/6.5/0.02/2\%,
            2.5/0.5/0.0/0\%, 2.5/1.5/0.16/16\%, 2.5/2.5/0.77/77\%, 2.5/3.5/0.03/3\%, 2.5/4.5/0.01/1\%, 2.5/5.5/0.01/1\%, 2.5/6.5/0.02/2\%,
            3.5/0.5/0.0/0\%, 3.5/1.5/0.06/6\%, 3.5/2.5/0.03/3\%, 3.5/3.5/0.82/82\%, 3.5/4.5/0.01/1\%, 3.5/5.5/0.01/1\%, 3.5/6.5/0.07/7\%,
            4.5/0.5/0.01/1\%, 4.5/1.5/0.01/1\%, 4.5/2.5/0.03/3\%, 4.5/3.5/0.01/1\%, 4.5/4.5/0.82/82\%, 4.5/5.5/0.03/3\%, 4.5/6.5/0.09/9\%,
            5.5/0.5/0.03/3\%, 5.5/1.5/0.10/10\%, 5.5/2.5/0.21/21\%, 5.5/3.5/0.01/1\%, 5.5/4.5/0.05/5\%, 5.5/5.5/0.53/53\%, 5.5/6.5/0.07/7\%,
            6.5/0.5/0.04/4\%, 6.5/1.5/0.01/1\%, 6.5/2.5/0.03/3\%, 6.5/3.5/0.01/1\%, 6.5/4.5/0.69/69\%, 6.5/5.5/0.02/2\%, 6.5/6.5/0.2/20\%
        } {
            \fill[cyan!30, opacity=0.7] (\x,\y) circle ({0.7*\size});
            \node[scale={max(0.5, 1.1*\size)}] at (\x,\y) {\text};
        }
        
        % Green Highlight for Low Quality Labels area
        % \fill[green!30, opacity=0.7] (0,0) rectangle (6,7);
        % \fill[green!30, opacity=0.7] (6,6) rectangle (7,7);
        \draw[red, very thick] (0,0) rectangle (6,7);
        \draw[red, very thick] (6,6) rectangle (7,7);
        \node at (3.5,-2) {a. Low Quality Labels (63K)};
    \end{scope}

    % Right Bubble Plot
    \begin{scope}[shift={(-11.0,0)}]
        % Grid and Axes
        \draw[gray!30, step=1] (0,0) grid (7,7);
        \draw[->] (0,0) -- (7.1,0) node[pos=0.5, below=0.5cm] {\normalsize Human Labels};
        \draw[->] (0,0) -- (0,7.1) node[pos=0.9, left=0.7cm, rotate=90] {\normalsize LLM Few-Shot Labels};

        % Axis Labels
        %\foreach \x in {1,...,7} {\node[below] at (\x-0.5,0) {\large \x};}
        %\foreach \y in {1,...,7} {\node[left] at (0,\y-0.5) {\large \y};}

        \foreach \x [count=\i from 1] in {A1,A2,A3,A4,A5,A6,A7} {
        \node[below] at (\i-0.5,0) {\small \x};
        }
        \foreach \y [count=\j from 1] in {A1,A2,A3,A4,A5,A6,A7} {
        \node[left] at (0,\j-0.5) {\small \y};
        }

        % Bubbles
        % Define bubble data - manually specify values
        \foreach \x/\y/\size/\text in {
            0.5/0.5/0.71/71\%, 0.5/1.5/0.02/2\%, 0.5/2.5/0.03/3\%, 0.5/3.5/0.01/1\%, 0.5/4.5/0.07/7\%, 0.5/5.5/0.03/3\%, 0.5/6.5/0.12/12\%,
            1.5/0.5/0.01/1\%, 1.5/1.5/0.43/43\%, 1.5/2.5/0.49/49\%, 1.5/3.5/0.02/2\%, 1.5/4.5/0.01/1\%, 1.5/5.5/0.02/2\%, 1.5/6.5/0.02/2\%,
            2.5/0.5/0.0/0\%, 2.5/1.5/0.16/16\%, 2.5/2.5/0.77/77\%, 2.5/3.5/0.03/3\%, 2.5/4.5/0.01/1\%, 2.5/5.5/0.01/1\%, 2.5/6.5/0.02/2\%,
            3.5/0.5/0.0/0\%, 3.5/1.5/0.06/6\%, 3.5/2.5/0.03/3\%, 3.5/3.5/0.82/82\%, 3.5/4.5/0.01/1\%, 3.5/5.5/0.01/1\%, 3.5/6.5/0.07/7\%,
            4.5/0.5/0.01/1\%, 4.5/1.5/0.01/1\%, 4.5/2.5/0.03/3\%, 4.5/3.5/0.01/1\%, 4.5/4.5/0.82/82\%, 4.5/5.5/0.03/3\%, 4.5/6.5/0.09/9\%,
            5.5/0.5/0.03/3\%, 5.5/1.5/0.10/10\%, 5.5/2.5/0.21/21\%, 5.5/3.5/0.01/1\%, 5.5/4.5/0.05/5\%, 5.5/5.5/0.53/53\%, 5.5/6.5/0.07/7\%,
            6.5/0.5/0.04/4\%, 6.5/1.5/0.01/1\%, 6.5/2.5/0.03/3\%, 6.5/3.5/0.01/1\%, 6.5/4.5/0.69/69\%, 6.5/5.5/0.02/2\%, 6.5/6.5/0.2/20\%
        } {
            \fill[cyan!30, opacity=0.7] (\x,\y) circle ({0.7*\size});
            \node[scale={max(0.5, 1.1*\size)}] at (\x,\y) {\text};
        }
        
        % Red Frame - diagonal line as shown in the image
        % \draw[red, thick] (0,0) -- (0,1);
        % \draw[red, thick] (0,0) -- (1,0);
        % \draw[red, thick] (1,0) -- (7,6);
        % \draw[red, thick] (0,1) -- (6,7);
        % \draw[red, thick] (7,6) -- (7,7);
        % \draw[red, thick] (6,7) -- (7,7);  
        \foreach \i in {0,...,6} {
          \draw[red, very thick] (\i,\i) rectangle ++(1,1);
        }
        % Volume text
        \node at (3.5,-2) {b. High Quality Labels (45K)};
    \end{scope}
    
    \end{tikzpicture}
    \caption{Two strategies for constructing the training set. The accidents used to create the training set are highlighted in red. (a) Assuming all human-labeled instances for types A1–A6 are correct, while for type A7, only labels agreed upon by both humans and the LLM few-shot approach are considered correct. Due to the weaker assumption in this approach, we name these 62K labels as low-quality. (b) Considering only labels where both humans and the LLM few-shot approach agree as correct, resulting in 45K high-quality labels.}
    \label{fig:bubble-plot-training-data}
\end{figure}
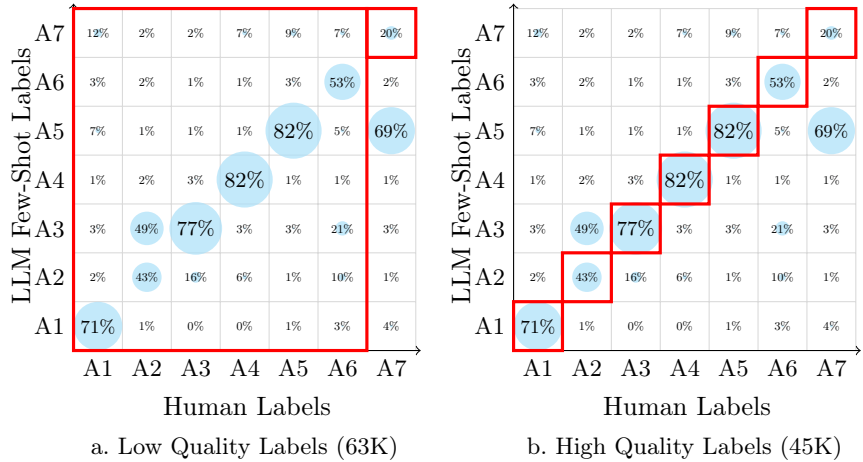

\paragraph{Training Set.}
One of the key motivations of this study is to detect mislabeled data. Therefore, relying solely on human labels is not ideal for this purpose. Instead, we explore two approaches to construct the training set, as illustrated in Figure \ref{fig:bubble-plot-training-data}. The main objective is to select accident labels that are more likely to be correct. To achieve this, we compare human labels with those generated by an LLM using a few-shot approach (explained in Section \ref{sec:few-shot}).

Our first strategy assumes that all human-labeled instances for accident types A1–A6 are correct, while for A7, only labels agreed upon by both humans and the LLM are considered valid. This approach results in approximately 62,000 training labels, which we refer to as presumably low-quality labels since the assumption about human labels being entirely correct is relatively weak. The second strategy is more conservative, considering only those labels where both the human annotators and the LLM agree. This produces a smaller but more reliable set of approximately 45,000 labels, which we refer to as presumably high-quality labels.

We train supervised models using both training sets and compare their performance to assess the impact of label quality on model accuracy.

\paragraph{Test Set.}
To ensure reliable evaluation of model performances, we ask domain specialists from the city of Munich to carefully label 236 traffic accidents. We use these expert-labeled examples as our test set to report results.

\subsubsection{Models \& Training Details}
\label{ssec:models}
\paragraph{MLP Model.}
Without using any text data, we employ a multi-layer perceptron (MLP) model to predict accident types based solely on tabular data. The model consists of feed-forward layers with skip connections and layer normalization. This model has 42 million parameters and is initialized randomly. The input is a fixed-size vector derived from the tabular features of the accident, and the model utilizes a softmax head to predict the accident type.

We train the MLP model on two Nvidia RTX A6000 GPUs. The optimization is performed using AdamW with an initial learning rate of 0.001, which decreases linearly by a factor of 0.9 every 5 epochs. We apply dropout with a probability of 0.1 and use a weight decay of 0.01 for regularization. The model is trained for 50 epochs, and we report the results from the best checkpoint based on the lowest validation loss.

\paragraph{LLM Few-shot Labeling.} 
We use the Gemma Few-shot labels for comparison with the other supervised trained models. The details of few-shot labeling are given in Section \ref{sec:few-shot}.

\paragraph{Finetuned XLM-R.}
To predict accident types using text data, we employ an encoder-only LLM. XLM-RoBERTa-Large (XLM-R), a multilingual model with 550 million parameters \cite{conneau2019unsupervised}, serves as our backbone. We use a pre-trained version of the model for transfer learning, fine-tuning it on our dataset. The model takes the accident text description as input and predicts the accident type through a softmax classification head with the model having 560 million parameters in total.

For fine-tuning XLM-R, we use Hugging Face \citep{wolf2019huggingface}. The model is fine-tuned on two Nvidia RTX A6000 GPUs. We use an initial learning rate of $5 \times 10^{-5}$, a weight decay of 0.01, and train for 6 epochs. The effective batch size is set to 128. As before, we select the best checkpoint based on the lowest validation loss.

\paragraph{Multimodal Model.}
To effectively handle both tabular and text data modalities, we design a multimodal model that integrates information from both sources. The architecture follows a two-branch structure:

\begin{itemize}
    \item \textbf{Textual Input Pathway:} The accident description is processed using the XLM-R model, which transforms the text into a text embedding via mean pooling.
    
    \item \textbf{Tabular Input Pathway:} The numerical and categorical accident-related features are fed into a multi-layer perceptron (MLP) model, which encodes them into a tabular feature embedding.
    
    \item \textbf{Fusion and Prediction:} The text embedding and tabular feature embedding are concatenated and passed through another MLP model, which acts as the final classifier with a softmax output layer to predict the accident type.
\end{itemize}
The XLM-R model is initialized with pretrained weights to leverage prior knowledge from multilingual text data, while the MLP components are randomly initialized. During training, all model parameters are updated.

The multimodal model is trained on two Nvidia RTX A6000 GPUs for 5 epochs with a batch size of 64. We use an initial learning rate of $1 \times 10^{-5}$, which decreases linearly by a factor of 0.7 every epoch. We apply a dropout rate of 0.2 and a weight decay of 0.01.

\setlength{\tabcolsep}{10pt} % Increase column spacing
\begin{table}[h]
\centering
\caption{Comparison of predictive models for accident classification using different data sources and corresponding model sizes.}
\label{tab:supervised-models}
\begin{tabular}{lll}
\toprule
Model & Data Modality & Parameter Size \\
\midrule
MLP & Tabular & 42 M \\
LLM Few-shot Labeling & Text & 27B \\
Finetuned XLM-R & Text & 560 M \\
Multimodal Model & Tabular + Text & 603 M \\
\bottomrule
\end{tabular}
\end{table}

\subsubsection{Model Comparisons}

To assess the different approaches and data modalities for accident classification, we compare the models using accuracy and weighted F1 score metrics calculated on the test set. Accuracy measures the proportion of correct predictions over all predictions. Weighted F1 score is computed as the weighted mean of F1 scores for each accident type
\begin{equation*}
\mathrm{Weighted\ }F_1\ \mathrm{Score}=\textstyle\sum_{i=1}^{7}w_i\cdot F_{1,i}
\end{equation*}
where \( F_{1,i} \) is the F1 score of class \( i \), calculated as the harmonic mean of precision and recall, and \( w_i = n_i /(\sum_{j=1}^{7} n_j)\)
with \( n_i \) the number of samples in class \( i \).
\subsection{Results}

\paragraph{Label Quality Effect.}

As discussed above, we constructed two training datasets: a larger one with lower-quality labels, containing approximately 62,000 accidents, and a smaller one with higher-quality labels, consisting of around 45,000 accidents. To examine the trade-off between dataset size and label quality, we trained three supervised models (MLP, Finetuned XLM-R, and Multimodal model) with the settings given in Section \ref{ssec:models}. We excluded the LLM few-shot approach from this comparison, as it does not involve parameter updates during training.

Table \ref{tab:predictive_model_performances} presents the performance of models trained on either low- or high-quality labels, evaluated on a test set of 236 expert-labeled accidents. Despite the smaller size of the high-quality label dataset, models trained on it achieve comparable performance to those trained on the larger low-quality label dataset. This highlights the importance of high-quality labels in model training. However, one should take into account the additional cost associated with generating high-quality labels—specifically, the computational expense of applying LLM few-shot labeling to a larger set of accidents. In our case, this was not an additional burden, as the few-shot labeling had already been applied to the full dataset for the preceding analysis.

\begin{table}[h]
    \centering
        \caption{Test set accuracy and weighted F1 scores for different models trained with datasets containing either low-quality (Low-Q) or high-quality (High-Q) labels. LLM Few-Shot Labeling is not explicitly trained, therefore presented in the center of both columns. Best accuracy and weighted F1 score values are highlighted in bold.}
    \renewcommand{\arraystretch}{1.3}
    \setlength{\tabcolsep}{2pt}
    \begin{tabular}{lccccc}
        \toprule
        \multirow{3}{*}{\textbf{Model}} & \multirow{3}{*}{\textbf{Data Modality}} & \multicolumn{4}{c}{\textbf{Test Set Performance}} \\
        \cmidrule(lr){3-6}
        & & \multicolumn{2}{c}{\textbf{Accuracy}} & \multicolumn{2}{c}{\textbf{W. F1 Score}} \\
        \cmidrule(lr){3-4} \cmidrule(lr){5-6}
        & & \textbf{Low-Q} & \textbf{High-Q} & \textbf{Low-Q} & \textbf{High-Q} \\
        \midrule
        MLP & Tabular & 0.53 & 0.53 & 0.49 & 0.48 \\
        LLM Few-Shot Labeling & Text & \multicolumn{2}{c}{0.61} & \multicolumn{2}{c}{0.58} \\
        Finetuned XLM-R & Text & 0.72 & 0.72 & 0.68 & \textbf{0.70} \\
        Multimodal Model & Text + Tabular & \textbf{0.73} & 0.70 & 0.68 & 0.68 \\
        \bottomrule
    \end{tabular}
\label{tab:predictive_model_performances}
\end{table}

\paragraph{Model Comparisons.}

Comparing the different models and modalities, the results indicate that the MLP model, which relies solely on tabular data, performs the worst among all approaches ($53\%$ accuracy, $0.49$ weighted F1 score). In contrast, all models incorporating text data---whether alone or in combination with tabular features---demonstrate better performance ($\geq 61\%$ accuracy, $\geq 0.58$ weighted F1 score). This suggests that textual information is the primary source of information for this classification task, whereas tabular data alone lacks the necessary detail for accurate accident classification.

When comparing the LLM Few-Shot approach ($61\%$ accuracy) to the finetuned XLM-R model ($72\%$ accuracy), we observe that finetuning leads to superior performance. Even though Gemma is a much larger model, with 27 billion parameters in a decoder-only architecture, it underperforms relative to the 560-million-parameter encoder-only XLM-R model. Two key factors likely contribute to this outcome. First, encoder-only models like XLM-R leverage a bidirectional attention mechanism, which is inherently more effective for text classification tasks compared to the autoregressive nature of decoder-only models. Second, finetuning allows the model to better adapt to domain-specific data, whereas few-shot prompting, even with six examples, does not provide the same level of task specialization.

Finally, comparing the Multimodal Model to the Finetuned XLM-R shows no large difference in performance when both text and tabular data are used. This suggests that textual data carries the most relevant information for accident classification in our predictive models.

\section{Conclusion and Future Directions}

In this study, we analyzed Munich traffic accidents using multiple data modalities to uncover meaningful patterns. Through semantic clustering, we identified distinct topics across seven accident categories, providing deeper insights into their characteristics.
To further investigate accident categorization, we employed an LLM with a few-shot approach and compared its results with human labels. Disagreements between the two revealed that many cases in the ``other accidents'' category involved damaged parked vehicles, supporting our findings from semantic clustering.

We also explored predictive modeling. Our results showed that models using text data outperformed those relying solely on tabular data, demonstrating the value of textual information for accident classification. Overall, our findings highlight the importance of NLP techniques in understanding traffic accidents. By leveraging textual data and machine learning, this approach offers valuable insights that can inform safety measures and contribute to the development of safer cities. 

Future developments could focus on improving and incorporating data-based classification into practice, for example by deploying such a model to make real-time suggestions to human labelers (human-in-the-loop) and use active learning approaches to improve model-based classification over time.

% Of course, authors have complete freedom on how they choose to structure their paper. Section headers from Introduction up to and including Conclusions are completely up to the discretion of the authors; use whichever structure you see fit. Title, Abstract, the credits environment, and References, however, are mandatory.

\begin{credits}
\subsubsection{\ackname} We thank the City of Munich for supporting this project and providing access to the dataset.
\end{credits}
\bibliographystyle{splncs04}
\bibliography{main}
%% Note that this preceding line implies that you store your BibTeX references in a file called 'mybibliography.bib'. If you instead store your references in a file with a different name, for instance 'references.bib', the preceding line should read '\bibliography{references}'. Whatever you do, DO NOT put the file name extension .bib inside the \bibliography command; this will trip up LaTeX compilers. 
%
\end{document}